\documentclass{article}
\usepackage{spconf,amsmath,graphicx}

\usepackage[moderate]{savetrees}
\usepackage{times}
\usepackage{latexsym}
\usepackage{multirow}
\usepackage{graphicx}
\usepackage{enumitem}
\usepackage[linesnumbered,ruled,vlined]{algorithm2e}
\usepackage{amsmath}
\usepackage{amssymb}
\usepackage{latexsym}
\usepackage[inkscapelatex=false]{svg}
\usepackage{url}

\SetKwInput{KwInput}{Input}                
\SetKwInput{KwOutput}{Output}              

\usepackage{amsthm,amsmath,amssymb}
\usepackage{mathrsfs}

\usepackage{makecell}
\usepackage{color}
\usepackage{booktabs}
\usepackage{hyperref}
\usepackage{float}
\usepackage[all]{nowidow}
\usepackage[T1]{fontenc}
\usepackage{microtype}
\usepackage{savetrees}

\title{IMPROVING MEDICAL DIALOGUE GENERATION WITH \\ ABSTRACT MEANING REPRESENTATIONS}
\name{Bohao Yang\textsuperscript{1*}, Chen Tang\textsuperscript{1,2*}, Chenghua Lin\textsuperscript{1$\dagger$}
\footnotemark[1]}
\address{\textsuperscript{1}Department of Computer Science, The University of Manchester, UK \\
\textsuperscript{2}Department of Computer Science, The University of Surrey, UK \\
\texttt{\{b.yang,c.lin\}@manchester.ac.uk}  \\
\texttt{\{chen.tang\}@surrey.ac.uk}
}

\begin{document}
%
\maketitle
%

\renewcommand{\thefootnote}{\fnsymbol{footnote}} 
\footnotetext[1]{Equal contribution.} 
\footnotetext[2]{Corresponding author.}

\begin{abstract}
Medical Dialogue Generation serves a critical role in telemedicine by facilitating the dissemination of medical expertise to patients. Existing studies focus on incorporating textual representations, which have limited their ability to represent the semantics of text, such as ignoring important medical entities. To enhance the model's understanding of the textual semantics and the medical knowledge including entities and relations, we introduce the use of Abstract Meaning Representations (AMR) to construct graphical representations that delineate the roles of language constituents and medical entities within the dialogues. In this paper, We propose a novel framework that models dialogues between patients and healthcare professionals using AMR graphs, where the neural networks incorporate textual and graphical knowledge with a dual attention mechanism. Experimental results show that our framework outperforms strong baseline models in medical dialogue generation, demonstrating the effectiveness of AMR graphs in enhancing the representations of medical knowledge and logical relationships. Furthermore, to support future research in this domain, we provide the corresponding source code at \textbf{\url{https://github.com/Bernard-Yang/MedDiaAMR.}}
\end{abstract}
\begin{keywords}
Abstract Meaning Representation, Dialogue Generation, Language Model, Artificial Intelligence, AMR Graph
\end{keywords}

\section{Introduction}
\label{sec:intro}

\begin{figure}[tb]
\centering
\includegraphics[width=0.75\linewidth]{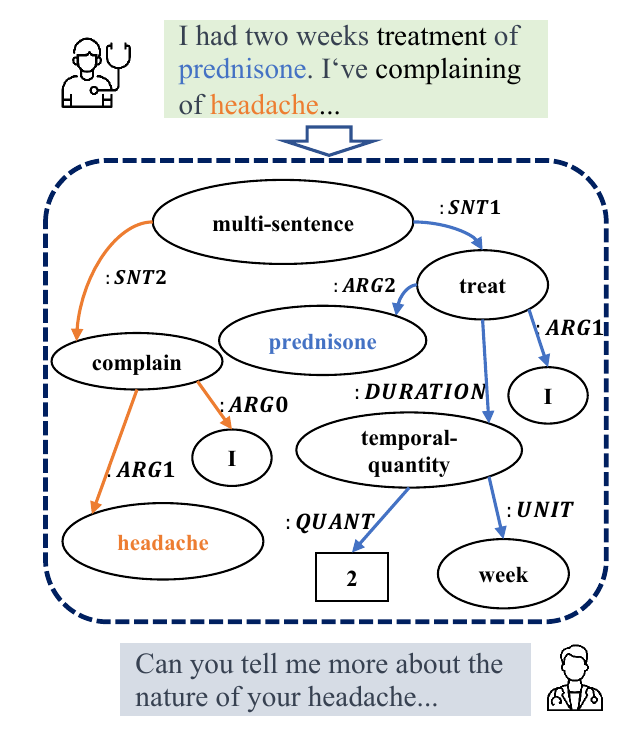}
\caption{An example showing how an AMR graph represents the dialogues of a patient. Terminologies from different sentences are shown in blue and orange colour.}
\label{fig:intro}
\end{figure}

The overarching goal of telemedicine is to provide patients with digital access to medical information, particularly in situations where direct access to a medical professional may be limited~\cite{tang2023terminology,abd2017analysing}. Prior research in this field has predominantly focused on incorporating medical knowledge by leveraging various types of additional annotations, including frequent items~\cite{givchi2022graph}, named entities~\cite{peng2021named}, entity relations~\cite{shang2011enhancing}, etc. However, unlike open-domain dialogue generation, the sequential features provided by text annotations struggle to comprehensively represent the intricate medical grounds and principles of diagnosis contained within medical dialogues.
To address this limitation, and to facilitate the incorporation of medical entities and their relations, our approach aims to construct dialogue-level Abstract Meaning Representation (AMR) graphs~\cite{bai-etal-2021-semantic}, and exploit both textual and graph-based features to enhance the language model on medical dialogue generation.

As shown in \autoref{fig:intro}, AMR graphs provide a structured and semantically rich representation of language~\cite{bai-etal-2021-semantic}. In the complex and critically important field of medicine, clear and precise communication is paramount. AMR graphs offer a standardised way to capture the relationships between words, entities, and their corresponding meanings, reducing ambiguity and potential misunderstandings in medical conversations. This enables medical professionals and patients to more easily interpret and trust the information conveyed, facilitating better decision-making, treatment adherence, and overall patient care. The incorporation of AMR graphs into the dialogue generation system enhances its capacity to comprehend the intricate semantics and contextual nuances implicitly embedded within textual content. Consequently, this integration empowers the system to generate context-aware medical dialogues. Our framework benefit the vanilla language models on achieving more precise and naturally articulated medical dialogue generation via incorporating the textual representations with the rich graph knowledge encapsulated by AMR graphs. 

In this study, we first construct AMR graphs by parsing the sentences within each patient's dialogue. Subsequently, these parsed AMR graphs are flattened and fed into a graph encoder, aligning with an independent sequence encoder for the text tokens in the input sentences. A module implemented by the dual-attention mechanism~\cite{bai-etal-2021-semantic}
is employed to incorporate the heterogeneous features originating from both the AMR graphs and input text. This combined representation is then used for the subsequent response decoding in an autoregressive manner. Our experimental results demonstrate that our approach substantially enhances the performance of the original language model and achieves the state-of-the-art performance, by capturing the additional structured knowledge during medical dialogue generation. Our contributions can be summarised as follows: (1) This is the first attempt to exploit AMR graphs for improving medical dialogue generation. (2) We propose a novel framework that incorporates both AMR graphs and text for medical dialogue generation and achieves the state-of-the-art performance; (3) We conduct comprehensive experiments to illustrate the effectiveness of our approach and provide a thorough analysis of the main components.

\vspace{0mm}
\section{Methodology}
\vspace{0mm}
\begin{figure*}[tb]
\centering
\includegraphics[width=0.90\linewidth]{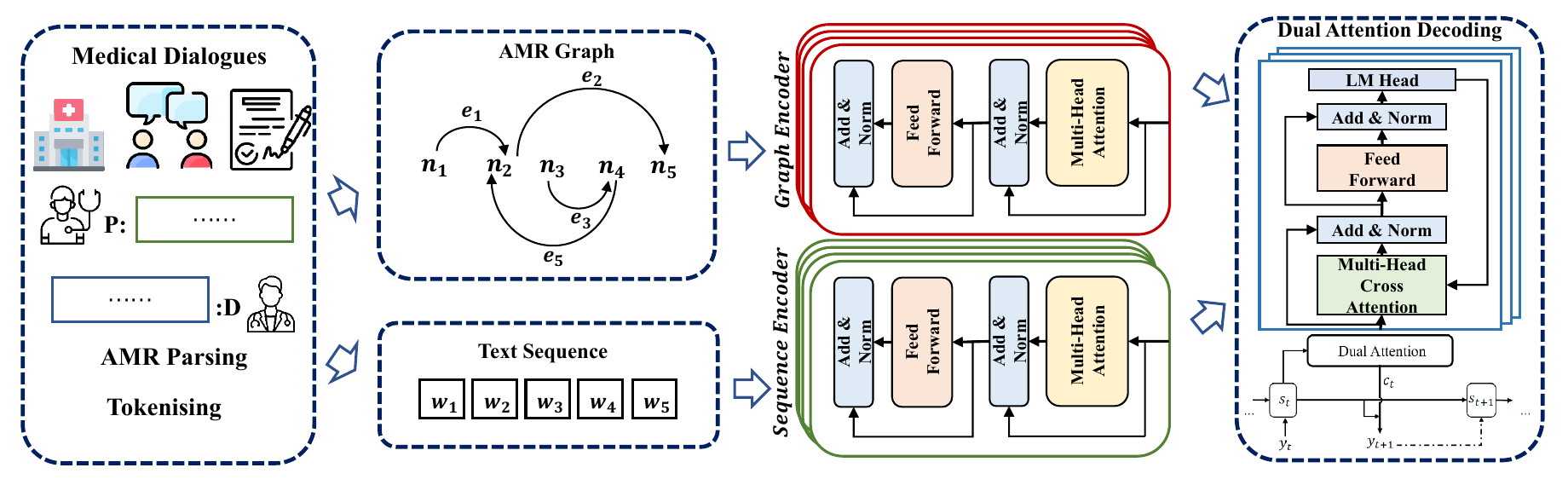}
\caption{The overview of our proposed framework.}
\label{fig:overview}
\end{figure*}


Our proposed framework is illustrated in \autoref{fig:overview}, which incorporates the heterogeneous features of the input text and parsed AMR graphs with two independent encoders. Subsequently, the decoder attends to the dual-attentioned features from encoders to autoregressively predict response tokens.

\subsection{Task Definition}
We define the task as follows: The input sequence data is
denoted as $X = {x_1, x_2, ..., x_n}$, which encompasses a medical inquiry along with the historical dialogue exchanges between the healthcare provider and the patient. 
In addition, the input AMR graphs of patient's inquiry is denoted as $G = {g_1, g_2, ..., g_n}$.
The primary objective of this task is to generate a response $Y = {y_1, y_2, ..., y_m}$, with the model operating under the premise of simulating the conditional probability distribution $P(Y|X, G)$. This formulation encapsulates the essence of our research endeavor, which revolves around generating doctor-like responses in a medical dialogue context.


\subsection{Sequence Encoding}

The Sequence encoder employed in this study adheres to the conventional Transformer architecturer~\cite{Vaswani2017AttentionIA}, which is designed to take the input patient's inquiry, denoted as $\mathbf{S}_i=$ $\left\{w_1, w_2, \ldots, w_\mathbf{S}\right\}$, and subsequently generates a corresponding sentence representation, denoted as $\mathbf{H}_{S}$. Formally, the sequence encoder is defined  as follows: 
\begin{align}
\mathbf{H}_{S}=\operatorname{Transformer}(\mathbf{S}) \\
h_i=\sum_{j=1}^{\mathbf{S}} \alpha_{i j}\left(W^H h_j\right) \\
\alpha_{i j}=\operatorname{Attention}\left(h_i, h_j\right)
\end{align}
where $\mathbf{H}_{S}=\left\{h_1, h_2, \ldots, h_\mathbf{S}\right\}$, $w_i$ represents  the $i$-th tokens, $\mathbf{S}$ signifies the sequence lengths,  $W^H$ represents learnable parameters. 

\subsection{Graph Encoding}
We employ a Graph Transformer~\cite{Zhu2019ModelingGS} to encode AMR graphs. An AMR graph $\mathbf{G}=\{\mathbf{N}, \mathbf{R}\}$ consists of graph nodes denoted by $\mathbf{N}$ and graph edges denoted by $\mathbf{R}$. Each edge $\mathbf{e} \in \mathbf{E}$ comprises of a set of elements $\left\{\mathbf{n_i}, \mathbf{r_{i j}}, \mathbf{n_j}\right\}$, symbolising the relation $\mathbf{r_{i j}}$ between two graph nodes $\mathbf{n_i}$ and $\mathbf{n_j}$.
The Graph encoder processes these nodes and relations as input and is formally defined as follows:
\begin{align}
\mathbf{H}_{G}=\operatorname{GraphEncoder}(\mathbf{N}, \mathbf{E}) \\
h'_i=\sum_{j=1}^M \hat{\alpha}_{i j}\left(W^V h'_j+W^R \boldsymbol{r}_{i j}\right)
\end{align}
where $\mathbf{H}_{G}=\left\{h_1', h_2', \ldots, h_M'\right\}$, and $W^V$ and $W^R$ are learnable parameters.

The graph attention of the Graph Transformer module is formally represented as:
\begin{align}
\hat{\alpha}_{i j} = &\frac{\exp \left(\hat{e}_{i j}\right)}{\sum_{m=1}^M \exp \left(\hat{e}_{i m}\right)} \nonumber \\ 
\hat{e}_{i j} = &\frac{\left(W^Q h'_i\right)^T\left(W^K h'_j+W^R \boldsymbol{r}_{i j}\right)}{\sqrt{d}}
\end{align}
where $W^Q$ and $W^K$ are  learnable parameters and $d$ is hidden state size.
The Graph Transformer effectively encodes structural information, represented by $\mathbf{r_{i j}}$, for all pairs of nodes within the AMR graphs. This incorporation of graph edge information enriches the node representations, enhancing the overall encoding process.

\subsection{Dual Attention Decoding}
Once we obtained the sentence representation $\mathbf{H}_{S}$ and graph representation $\mathbf{H}_{G}$, we proceed to input them into the decoder equipped with a dual attention mechanism.
For each decoder hidden state $d_t$, the dual-attention mechanism produce a sequence context vector $c_{t S}$ and a graph context vector $c_{t G}$ at each time step $t$ :

\begin{align}
c_{t i}=\sum_{i=1}^S \hat{\alpha}_{t i} h_i \\
\hat{\alpha}_{t i}=\operatorname{Attention}\left(d_t, h_i\right) \\
c_{t j}=\sum_{j=1}^M \hat{\alpha}_{t j} h'_j\\
\hat{\alpha}_{t j}=\operatorname{Attention}\left(d_t, h'_j\right)
\end{align}
Subsequently, we concatenate the sequence context vector $c_{t S}$ and the graph context vector $c_{t G}$ to compose the ultimate context vector $\hat{c}_t$. This combination is formally represented as:
\begin{align}
c_t=W^C\left[c_{t S} ; c_{t G}\right]+b
\end{align}
where $W^C$ and $b$ are learnable parameters.


\subsection{Training and Inference}
Finally, those fused sentence and graph features are autoregressively decoded to predict responses, where the predicted tokens are forced to be close to the golden responses. We train the whole model with the loss function as follows:
\begin{align}
\mathcal{L}=-\frac{1}{N} \sum_{n=1}^N \log P(Y \mid X, G)
\end{align}
where $N$ denotes the size of the training data, and $\mathcal{L}$ is the cross entropy between the predicted response tokens and the golden responses.

\vspace{0mm}
\section{Experiment}
\vspace{0mm}

\begin{table*}[ht]
\centering \small
\resizebox{0.90\linewidth}{!}{
\begin{tabular}{l|cccc|ccc|cccc}
\toprule
\textbf{Model} & \textbf{B-1$\uparrow$} & \textbf{B-2$\uparrow$} & \textbf{B-3$\uparrow$} & \textbf{B-4$\uparrow$} & \textbf{R-1$\uparrow$} & \textbf{R-2$\uparrow$} & \textbf{R-L$\uparrow$} & \textbf{Dist-1$\uparrow$} & \textbf{Dist-2$\uparrow$} & \textbf{Dist-3$\uparrow$} & \textbf{Dist-4$\uparrow$} \\
\hline
    \textbf{GPT-2}  & 0.0725 & 0.0376 & 0.0267 & 0.0218 & 12.5431 & 4.3497 & 9.8125 & 0.0048 & 0.0245 & 0.0511 & 0.0725 \\
    \textbf{DialoGPT} & 0.0599 & 0.0310 & 0.0225 & 0.0187 & 9.9041 & 3.5320 & 8.0158 & 0.0036 & 0.0169 & 0.0354 & 0.0531 \\
    \textbf{T5-base} & 0.1255 & 0.0585 & 0.0319 & 0.0188 & 12.7913 & 2.1351 & 9.8115 & 0.0026 & 0.0102 & 0.0183 & 0.0265 \\
    \textbf{T5-large} & 0.0952 & 0.0529 & 0.0388 & 0.0321 & 16.1452 & 5.8471 & 12.3637 & 0.0055 & 0.0282 & 0.0593 & 0.0883 \\
    \textbf{BART-base} & 0.1205 & 0.0631 & 0.0423 & 0.0321 & 19.3620 & 5.1807 & 11.4057 & 0.0046 & 0.0374 & 0.1125 & 0.2087 \\
    \textbf{BART-large} & 0.1142 & 0.0621 & 0.0420 & 0.0319 & 19.4735 & 5.3347 & 11.4324 & 0.0055 & 0.0435 & 0.1131 & 0.1933 \\ 
    \textbf{Terms-BART} & 0.1547 & 0.0822 & 0.0555 & 0.0421 & 20.2111 & 5.4167 & 13.0137 & 0.0071 & \textbf{0.0453} & \textbf{0.1462} & 0.2899 \\
    \hline
    \textbf{Ours} &   \textbf{0.1705} & \textbf{0.1411} & \textbf{0.1613} & \textbf{0.1336} & \textbf{35.7853} & \textbf{14.5550} & \textbf{23.9057} & \textbf{0.0088} & 0.0179 & 0.1376 & \textbf{0.4321} \\
    \textbf{- w/o text} & 0.1222 & 0.0716 & 0.0620 & 0.0336 & 29.8947 & 9.1014 & 18.4940 & 0.0062 & 0.0082 & 0.0448 & 0.1626 \\ 
    \textbf{- w/o AMR} & 0.0802 & 0.0432 & 0.0304 & 0.0221 & 24.1742 & 5.4295 & 12.9662 & 0.0042 & 0.0059 & 0.0376 & 0.0904 \\ 
\bottomrule
\end{tabular}
}
\caption{\label{auto evaluation}
The automatic evaluation compares the performance of our model with various baseline systems on medical dialogue generation. The best-performing model for each metric is highlighted in bold. 
}
\end{table*}

\vspace{0mm}
\subsection{Experimental Setup}
\vspace{0mm}
\noindent\textbf{Data Preparation.}~~In order to prepare the input data for the Graph Transformer, We adopt an open-source pre-trined AMR parser~\cite{cai-lam-2020-amr} in transforming utterances into corresponding AMR.
Subsequently, we streamline the acquired AMR graphs through the utilization of the AMR simplifier~\cite{Konstas2017NeuralAS}. This simplification process allows us to extract pertinent concepts and relationships, which are pivotal components for our subsequent analytical endeavors.

\noindent\textbf{Baselines.}~~We conduct a comprehensive comparative analysis of our proposed framework against the competitive language models used in several recent advances~\footnote{Our approach can also be applied to larger language models, e.g. ChatGPT, but to be fair, some extremely large language models (ChatGPT has around 50 times parameter size than our baselines.) are not included in comparisons.} \cite{zeng-etal-2020-meddialog,zhou-etal-2021-generation,wang-etal-2021-fast,tang-etal-2023-enhancing}, as well as those relevant to our task~\cite{tang-etal-2022-etrica}. The models under consideration are as follows:
BART \cite{lewis2019bart}: A widely used language model renowned for  text generation tasks.
T5 \cite{raffel2020exploring}: A language model distinguished by its encoder-decoder architecture, successfully adapted for diverse generation tasks.
GPT-2 \cite{radford2019language}: A popular pre-trained language model widely applied in dialogue generation tasks.
DialoGPT \cite{zhang2019dialogpt}: A dialogue-oriented pre-trained GPT-2 model known for its strong performance in dialogue generation tasks.
Term-BART \cite{zhang2019dialogpt}: An advanced framework specifically tailored for medical dialogue generation, representing the state-of-the-art in the field.

\noindent\textbf{Metrics.} To comprehensively assess the efficacy of our proposed framework, we employ a range of both referenced and unreferenced metrics in our experiments. BLEU (B-$n$) \cite{Bleu} and ROUGE (R-$n$) \cite{lin2004rouge}, including the longest common subsequence variant (R-L), gauge the quality of generated responses by assessing their $n$-gram overlaps with reference responses. Additionally, we employ the metric of response diversity, following the approach outlined in \cite{li2015diversity}, by quantifying the $n$-gram distinction, denoted as Diversity (D-$n$), within generated responses, where $n$ signifies the $n$-gram order. This comprehensive set of metrics ensures a rigorous evaluation of our framework's performance and its comparison to the baseline models.

\subsection{Implementation Details}
The pre-trained models for baselines employed in this study originate from publicly accessible checkpoints hosted on the Hugging Face platform\footnote{\url{https://huggingface.co/models}}. All the models undergo a training process spanning a maximum of $10$ epochs, executed on a Tesla A100 computational unit, over a duration of approximately one full day.
Our training configuration prescribes a batch size of $36$, with a learning rate of $1e^{-4}$, and employs the Adam optimizer for the training procedure. These meticulous specifications underpin the rigorous methodology adopted in our academic inquiry.

\vspace{0mm}
\subsection{Experimental Results}
\label{sec:exp}
\vspace{0mm}
As shown in Table \ref{auto evaluation}, our proposed framework 
outperforms all baseline models across both referenced and unreferenced evaluation metrics. This outcome underscores the positive effect of introducing Abstract Meaning Representation (AMR) on enhancing the knowledge incorporating capabilities of the language model. Specifically, the marked enhancement in referenced metrics, such as BLEU and ROUGE, indicates the model's capacity to generate responses that encapsulate medical knowledge grounds as the gold references do. 
In addition, the substantial improvement on the unreferenced metrics, i.e. Dist-1 and Dist-4, signifies that the diversity of generated text is also improved via the incorporation of the graph representations. We posit that this outcome can be attributed to the improved flexibility inherent in the graph representations of the dialogue context. It is noticeable that the baselines of BART have better Dist-1 and Dist-3 is because they have more repeated short phrases generated in the dialogues, e.g. ``I am''.. 


Furthermore, the experiments also investigate the impact of model size, ranging from the base to the large variant. While it is intuitive to expect performance improvements with larger models, the empirical results on various metrics reveal that models with larger parameters, such as BART-large and T5-large, do not consistently outperform their smaller counterparts, BART-base and T5-base. This suggests that merely increasing the model size does not guarantee performance enhancements. It becomes apparent that existing language models, which primarily encode the superficial structure of dialogues, face substantial challenges in incorporating domain-specific knowledge without explicitly modeling medical concepts and relationships. This indicates that our utilisation of AMR graphs serves as a crucial bridge to address this gap in medical concept representations. 

In the ablation study, the results further demonstrates the effectiveness
of introducing AMR graphs as the additional feature to the language model. our proposed model, which encodes the heterogeneous features of both text and abstract meaning representations in a graph format, demonstrates significant improvements across all evaluated metrics.  Notably, several metrics exhibit over two-fold increments, such as B-4, which experiences a substantial rise from approximately 0.0221 to 0.1336, and Dist-4, which increases from approximately 0.09 to 0.43. Another finding is certain metrics (i.e. BLEU) indicate that our proposed model achieves scores surpassing the mathematical sum of ablated models (" -w/o text" and " -w/o AMR"). This implies that these two features encapsulate distinct semantic aspects of the dialogues. Therefore, the integration of both graphical and textual features profoundly enhances the process of medical dialogue generation.

\begin{table}[ht]
\centering
\resizebox{\linewidth}{!}{
\begin{tabular}{{p{1.2\linewidth}}}
\toprule[2pt]
\textbf{Context:} Patient: I have a \textbf{sharp pain} at the front of my \textbf{hip} at the top of my \textbf{thigh} where it nears my \textbf{lower abdomen}. This \textbf{pain} only occurs when i bend over and lasts there after for a few seconds. It then goes away and does not recur until i bend over again. Any idea what this could be? maybe a hernia?\\
\midrule[1pt]
\textbf{Reference:} Hello, it is unlikely be an hernia. The \textbf{symptoms} are more likely suggestive of \textbf{musculo-skeletal causes} like \textbf{sprain} or contusion. As of now you can] use \textbf{analgesics/anti inflammatory combination} like \textbf{aceclofenac/seratiopeptdase} for \textbf{symptomatic relief}. You can go for an \textbf{ultrasound scan} to rule out \textbf{hernia}. Hope i have answered your query. Let me know if i can assist you further. Wishing you all the best. Regards, dr. shinas hussain, general \& family physician. \\
\midrule[1pt]
\textbf{Terms-BART:}  Hello and welcome to ‘ask a doctor’ service. I have reviewed your query and here is my advice. As per your case history of hip \textbf{pain}, it could be due to \textbf{hip muscle spasm}. For this reason, I would recommend consulting with your attending \textbf{physician} for a \textbf{physical exam} and some \textbf{tests}. Hope I have answered your query. Let me know if I can assist you further. Regards, Dr. Praveen Tayal, orthopaedic surgeon apollo hospitals.\\
\midrule
\textbf{Ours:} Hello, it is unlikely be an \textbf{hernia}. You can use \textbf{analgesics/anti inflammatory} combination like \textbf{aceclofenac/seratiopeptdase} for symptomatic relief. Regards, dr. shinas hussain, general \& family physician. \\
\midrule[1pt]
\textbf{- w/o text: }Hi, you can take  drugs like \textbf{ibuprofen} for pain relief. You can take for a \textbf{mri abdomen}. Hope i have answered your query. Let me know if i can assist you further. dr. shinas hussain, \& family physician.
\\
\midrule
\textbf{- w/o AMR: }Hi, thanks is a to due infection of pain. You have take \textbf{ibuprofen} for a relief. You can take for a \textbf{mri}. Hope i have answered your query. Let me know if i can assist you further. Regards, dr. shinas hussain, general \& family physician.\\ 
\bottomrule[2pt]
\end{tabular}
}
\caption{A Sample collected for the case study, where terminologies are highlighed in \textbf{bold}.}
\label{tab:case study1}
\end{table}


For qualitative analysis, we present the generated dialogues from different models in \autoref{tab:case study1}. It can be observed that even though the previous state-of-the-art model \textbf{Terms-BART} uses more terminological knowledge when answering the question, dialogues generated by \textbf{ours} address the patient's queries more effectively, where the terminologies in the response are more related to the context. This improvement can be attributed to our model's enhanced capability in capturing semantics and logical structures, which are attained from the additional Abstract Meaning Representation (AMR) graphs.The performance improvement of our model is further demonstrated via a comparison with the response generated by the model variant that does not utilise AMR representations (\textbf{- w/o AMR}).

\vspace{0mm}
\section{Conclusion}
\vspace{0mm}
In conclusion, our approach, which are the first to integrates Abstract Meaning Representations (AMR) to capture the semantics and medical terminologies embedded within dialogues, offers a novel and effective framework for modeling patient-doctor dialogues. Through incorporating the textual and graphical knowledge into a unified language model, our proposed framework achieves the state-of-the-art performance, and the experimental results show a substantial performance improvement over baseline models in the domain of medical dialogue generation. This demonstrates the strong potential of our framework in empowering language models to leverage medical knowledge and logical relationships, ultimately enhancing the quality of medical dialogue generation.


\bibliographystyle{IEEEbib}
\bibliography{strings,refs}

\end{document}